\documentclass{article}
\usepackage{spconf,amsmath,amssymb,graphicx,hyperref,subcaption}
\usepackage{array}
\usepackage{makecell}
\usepackage{booktabs}
\usepackage[table]{xcolor}
\usepackage{multirow}
\usepackage{array}
\usepackage{flafter}
\usepackage{float}
\usepackage{tikz}
\hypersetup{hidelinks}
\usetikzlibrary{arrows.meta}
\providecommand{\rvec}{\boldsymbol{r}}
\usepackage[T1]{fontenc}

\newcommand{\qwen}{Qwen2-Audio}

\usepackage{pifont}
\newcommand{\cmark}{\ding{51}}
\newcommand{\xmark}{\textcolor{gray!60}{\ding{55}}}

\newcommand{\s}[1]{\,{\tiny$\pm#1$}}
\usepackage{caption}
\providecommand{\baselinesteersep}{%
  \noalign{\vskip 1pt \color{black!30}%
    \hrule height 0.25pt \vskip 1.5pt \hrule height 0.25pt%
    \color{black}\vskip 3pt}%
}

\title{Steering Speech-Language Models: Training-Free Task Specialization \\ via  Contrastive Activation Addition}

\name{\begin{tabular}{@{}c@{}}
      Severin Baroudi$^{1,2}$ \quad Yanis Labrak$^{3}$ \quad Pierfrancesco Melucci$^{2,4}$ \quad Sergio Burdisso$^{3}$  \\ Petr Motlicek$^{3,5}$ \quad
      Herv\'e Bredin$^{6}$ \quad Mirco Ravanelli$^{2,7}$ \quad Ricard Marxer$^{1,8}$
      \end{tabular}}
\address{
  $^{1}$Univ Toulon, Aix Marseille Univ, LIS, CNRS, France \quad
  $^{2}$Mila -- Quebec AI Institute \\
  $^{3}$Idiap Research Institute, Martigny, Switzerland
  $^{4}$Sapienza University of Rome, Italy \\
  $^{5}$Brno University of Technology, Brno', Czech Republic \quad
  $^{6}$pyannoteAI, Toulouse, France \\
  $^{7}$Concordia University, Montr\'eal, Canada \quad
  $^{8}$CNRS, ILLS, Montr\'eal, Canada}

\begin{document}
\ninept
\maketitle

\begin{abstract}
Activation steering has proven effective for controlling the behavior of Large Language Models (LLMs) at inference time, but its application to SpeechLLMs remains new, and training-free steering approaches for such models are still largely unexplored. We propose a training-free Contrastive Activation Addition (CAA) protocol that derives steering vectors for common speech tasks  (e.g. transcription) in SpeechLLMs from a small number of labeled utterances. We showcase that adding these vectors in the representation space, at inference time, enforces better the targeted speech task. We further show that, when combined with prompting, these vectors yield to consistent improvement over prompting alone on most evaluated tasks such as Automatic Speech Recognition (ASR) or Emotion Recognition (ER), and transfer to out-of-domain data. We additionally demonstrate the usefulness of script-normalization directions to enforce the target script of a specific language.
\end{abstract}

\section{Introduction}
\label{sec:intro}

The field of mechanistic interpretability provides a set of general-purpose
tools, from linear probing to causal intervention on
intermediate activations~\cite{rome}, and sparse dictionary
learning~\cite{sae}, with the objective of isolating human-interpretable
components on the internal representations of a neural network. Among these techniques, activation steering directly alters the internal hidden
states of a given Large Language Model (LLM), during generation, to control its
behavior, style, tone, or persona without any retraining of the
model~\cite{caa}. In particular, these tools have established that humanly interpretable attributes are encoded in LLMs in a linear fashion~\cite{linearhyp} via specific vectors. With the advantage of being easily interpretable and manipulable (as opposed to sparse autoencoders~\cite{sae}), the use of steering vectors has recently been transposed and adapted to other modalities, with the goals of improving the controllability, safety, and performance of language models involved in images and audio. 


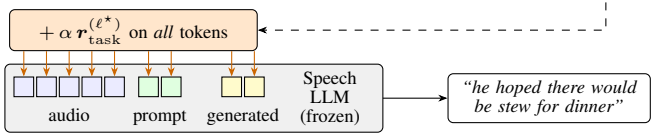
\begin{figure}[t]
\centering
\begin{tikzpicture}[
  font=\scriptsize,
  >={Stealth[length=3.5pt]},
  box/.style={draw, rounded corners=2pt, align=center, inner sep=2.5pt},
  aud/.style={box, fill=blue!8},
  posb/.style={box, fill=green!12, minimum width=19mm},
  negb/.style={box, fill=red!10, minimum width=19mm},
  llm/.style={box, fill=gray!12},
  vec/.style={box, fill=orange!20},
  op/.style={draw, circle, inner sep=0.5pt, font=\footnotesize},
  tok/.style={draw, minimum width=2.6mm, minimum height=2.6mm, inner sep=0},
  toklab/.style={anchor=base, inner sep=0},
  lab/.style={font=\footnotesize\bfseries, anchor=west, inner sep=0},
]
\node[lab] at (0,2.35) {(a) Extraction of a task direction};
\node[aud, minimum height=14mm] (A) at (0.40,1.2) {audio\\pool\\$\mathcal{A}$};
\node[posb] (P) at (2.23,1.6) {$\mathcal{D}_{\mathrm{pos}}$: ``Transcribe\ldots''};
\node[negb] (N) at (2.23,0.8) {$\mathcal{D}_{\mathrm{neg}}$: ``Summarize\ldots''};
\node[llm, minimum height=14mm, minimum width=13mm] (M) at (4.35,1.2) {Speech\\LLM\\(frozen)};
\node (mp) at (5.98,1.6) {$\boldsymbol{\mu}^{(\ell)}(\mathcal{D}_{\mathrm{pos}})$};
\node (mn) at (5.98,0.8) {$\boldsymbol{\mu}^{(\ell)}(\mathcal{D}_{\mathrm{neg}})$};
\node[op] (minus) at (7.05,1.2) {$-$};
\node[vec] (R) at (7.85,1.2) {$\rvec^{(\ell)}_{\mathrm{task}}$};
\draw[->] (A.east|-P) -- (P);
\draw[->] (A.east|-N) -- (N);
\draw[->] (P) -- (M.west|-P);
\draw[->] (N) -- (M.west|-N);
\draw[->] (M.east|-mp) -- (mp);
\draw[->] (M.east|-mn) -- (mn);
\draw[->] (mp.east) -| (minus.north);
\draw[->] (mn.east) -| (minus.south);
\draw[->] (minus) -- (R);
\node[font=\scriptsize, align=center, text=black!70] at (5.95,0.15) {mean last-token activation\\at layer $\ell$, Eq.~\eqref{eq:mean}};
\node[lab] at (0,-0.1) {(b) Steering at inference};
\draw[rounded corners=2pt, fill=gray!12] (-0.1,-2.05) rectangle (4.9,-1.15);
\node[align=center] at (4.2,-1.6) {Speech\\LLM\\(frozen)};
\foreach \i in {0,...,4} {\node[tok, fill=blue!8]   (a\i) at (0.15+\i*0.3,-1.45) {};}
\foreach \i in {0,...,1} {\node[tok, fill=green!12]  (p\i) at (1.8+\i*0.3,-1.45) {};}
\foreach \i in {0,...,1} {\node[tok, fill=yellow!25] (g\i) at (2.9+\i*0.3,-1.45) {};}
\node[toklab] at (0.75,-1.9) {audio};
\node[toklab] at (1.95,-1.9) {prompt};
\node[toklab] at (3.05,-1.9) {generated};
\node[vec, text width=31mm] (ar) at (1.6,-0.7)
  {$+\,\alpha\,\rvec^{(\ell^{\star})}_{\mathrm{task}}$ on \emph{all} tokens};
\foreach \t in {a0,a1,a2,a3,a4,p0,p1,g0,g1}
  {\draw[->, orange!80!black] (\t.north|-ar.south) -- (\t.north);}
\node[box, text width=25mm] (out) at (7.1,-1.6)
  {\textit{``he hoped there would be stew for dinner''}};
\draw[->] (4.9,-1.6) -- (out.west);
\draw[->, dashed] (R.south) |- (ar.east);
\end{tikzpicture}
\caption{Method overview: (a) extraction of a task direction (ASR in this case) from contrasting instructions on the same audio; (b) steering at inference: the direction is added at a single layer $\ell^{\star}$ to every token, audio positions included.}
\label{fig:method}
\end{figure}

For speech, few activation steering approaches involving directions have been
proposed. In speech synthesis, a difference in means between emotional and neutral activations controls the emotion of the generated speech~\cite{emosteer}, and activation editing corrects prosody and mispronunciation post hoc~\cite{cae}. These interventions effectively control a paralinguistic attribute or a generation style. The framework SALSA~\cite{salsa} freezes the backbone of a SpeechLLM and trains neural networks to learn steering vectors to inject at inference, by maximizing the likelihood of the reference transcript given the audio, making the intervention a parameter-efficient alternative to
fine-tuning on low-resource speech. For SpeechLLMs, only two recent training-free methods obtain such vectors using Contrastive Activation Addition (CAA) in the latent space. The first~\cite{accentsteer} improves transcription by identifying accent-related directions from pairs of utterances that share the same linguistic content but differ in accent, while the second~\cite{nudging} extracts a chain-of-thought direction to strengthen reasoning. The latter suggests that performing this type of intervention, improves the ability of the model to reason and follow instructions.


In practice, SpeechLLMs are specialized through textual prompting, whose performance can vary substantially across semantically equivalent instructions, a phenomenon known as \emph{prompt sensitivity}~\cite{sclar2024quantifying}. Supervised fine-tuning (SFT) removes this dependency, but requires labeled data and a new set of weights per task, which make the deployment difficult. Training-free CAA offers a third route, yet in SpeechLLMs it has only been applied to narrow settings (accented transcription or chain-of-thought reasoning), and whether it can specialize a single SpeechLLM towards a range of speech tasks remains open. This motivates our central research question:
\textbf{Are speech tasks linearly encoded in the representation space of SpeechLLMs ?} If so, can task directions be leveraged to   
\textbf{\textit{(a)}} serve as an alternative to prompting for audio tasks (e.g.\ transcription),
\textbf{\textit{(b)}} complement textual prompting and improve controllability, therefore reducing \textit{prompt sensitivity} ? Finally, \textbf{\textit{(c)}} are these task directions generalizable and dataset-agnostic ?



In this paper, we propose a generalizable approach involving CAA for inducing and controlling speech tasks in SpeechLLMs.
\textbf{(i)}~We introduce a simple training-free procedure that extracts a task-specific steering vector from a small set of speech or audio examples. By injecting this direction at inference time, we show that a SpeechLLM can be specialized towards a specific speech task without any textual instruction, and that combining steering with prompting outperforms prompting alone while reducing its sensitivity to the choice of prompt.
\textbf{(ii)}~Moreover, we demonstrate that these task steering vectors are localized and dataset-agnostic, allowing transferability across out-of-domain corpora, such as Automatic Speech Recognition (ASR).
\textbf{(iii)}~Finally, we show that the textual alphabet of the SpeechLLM can be controlled with the same injection technique, to enforce a non-Latin script where the model would otherwise default to Latin characters, allowing for a more faithful transcription.

\section{Activation Addition}
\label{sec:method}

\subsection{Extracting a task direction}
\label{ssec:extraction}

We display the complete steering protocol in Figure  \ref{fig:method}. Let $\mathcal{M}$ be a SpeechLLM whose language backbone is a stack of $L$
transformer blocks operating on a residual stream of width $d$. Given an audio
clip $a$ and a textual instruction $p$, the input sequence is the concatenation
of the $T_a$ audio tokens produced by the audio encoder and the $T_p$ tokens
of the instruction. We write $\boldsymbol{h}^{(\ell)}_{i}(a,p)\in\mathbb{R}^{d}$
for the residual activation at site $\ell$ and position $i$, where
$\ell\in\{0,\dots,L\}$ indexes the embedding output and the $L$ block outputs.

Following the refusal-direction protocol~\cite{refusal}, we read activations at
the \emph{post-instruction} position $i=-1$, i.e.\ the last token of the prompt,
which has been shown to aggregate the instruction semantics in textual LLMs.
As \cite{refusal} demonstrated that a refusal direction can be identified from a contrast between the residual streams of refused and complied-with requests, we assume that a task-conditioned behavior can be recoverable from a contrast between different instructions given to a SpeechLLM. We therefore define
a positive set $\mathcal{D}_{\mathrm{pos}}$ of prompts carrying a target
task (e.g.\ \textit{``transcribe the given speech''}, \textit{``give the
verbatim transcript of the given audio''}) and a negative set
$\mathcal{D}_{\mathrm{neg}}$ of instructions for other tasks (e.g.\
\textit{``summarize the given speech''}, \textit{``translate the given
speech''}). Both sets are paired with the same pool $\mathcal{A}$ of audio clips,
so that the mean activation of each set is:
\begin{equation}
  \boldsymbol{\mu}^{(\ell)}(\mathcal{D}) =
  \frac{1}{|\mathcal{A}|\,|\mathcal{D}|}
  \sum_{a\in\mathcal{A}}\;\sum_{p\in\mathcal{D}}
  \boldsymbol{h}^{(\ell)}_{-1}(a,p),
  \label{eq:mean}
\end{equation}
and the task direction is defined as the difference in means:
\begin{equation}
  \boldsymbol{r}^{(\ell)}_{\mathrm{task}} =
  \boldsymbol{\mu}^{(\ell)}(\mathcal{D}_{\mathrm{pos}}) -
  \boldsymbol{\mu}^{(\ell)}(\mathcal{D}_{\mathrm{neg}}).
  \label{eq:direction}
\end{equation}
Since the two sets share the same audio pool $\mathcal{A}$, the acoustic, speaker and lexical contributions are assumed to cancel on average, so that only the instruction-induced component remains.

\subsection{Task induction via activation addition}
\label{ssec:induction}

In CAA, steering vectors are traditionally applied at a single site $\ell^{\star}$, following applications in textual LLMs \cite{caa, refusal}. At every decoding steps, the
targeted residual activation is shifted by:
\begin{equation}
  \tilde{\boldsymbol{h}}^{(\ell^{\star})}_{i} =
  \boldsymbol{h}^{(\ell^{\star})}_{i} +
  \alpha\,\boldsymbol{r}^{(\ell^{\star})}_{\mathrm{task}},
  \qquad i \in \mathcal{P},
  \label{eq:addition}
\end{equation}
where $\alpha$ is a task-specific strength expressed in units of
$\lVert\boldsymbol{r}^{(\ell^{\star})}_{\mathrm{task}}\rVert$ and $\mathcal{P}$ is
the set of steered token positions. For textual LLMs, \cite{refusal} assumes that induction can be performed on all token positions (including the input prompt tokens).
In SpeechLLMs, current approaches usually process audio using a speech encoder architecture \cite{whisper, qwen2audio, phi4mm} and feed the representations into a subsequent language backbone, as continuous embeddings (or discrete tokens) part of the input prompt alongside text tokens. As a result, steering \emph{all} token positions (including the audio tokens) could potentially blur or destroy the audio content processed by the SpeechLLM. We therefore consider two variants: \textit{all-position} steering
($\mathcal{P}$ = audio $\cup$ input tokens $\cup$ generated tokens) and \textit{text-position} only
steering, which leaves the audio frames untouched ($\mathcal{P}$ = input tokens $\cup$ generated tokens), reported in Table~\ref{tab:taskonly_textonly}.
The strongest steering direction for a given task is defined empirically as the right combination of amplitude $\alpha$ and layer $\ell^{\star}$, that optimizes the target metric (i.e.\ minimizes Word Error Rate (WER) for ASR and maximizes Weighted Accuracy (WA) for ER). It is selected via a grid search performed using a small audio pool $\mathcal{A}$, drawn exclusively from the training sets of the targeted task.



Because Eq.~\eqref{eq:addition} is linear, several directions can be injected at
once, provided that they remain mostly orthogonal in the representation space. In this paper, we study the joint injection of a script-normalization direction and a transcription direction, in order to enforce a target alphabet when ASR must be performed in non-latin languages. The script-normalization direction is derived textually by asking a SpeechLLM to translate a sentence from English to one of our targeted transcription languages, also providing words from the target alphabet (e.g. \textit{Translate the following English text into Arabic script} (\raisebox{-0.3ex}{\includegraphics[height=1.9ex]{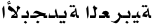}})\footnote{Translates to: ``The Arabic alphabet.''}: \textit{I would like a cup of coffee}). The latter is applied in conjunction with transcription steering as follows:
\begin{equation}
  \tilde{\boldsymbol{h}}_{i} = \boldsymbol{h}_{i} +
  \alpha\,\boldsymbol{r}_{\mathrm{task}} + \beta\,\boldsymbol{r}_{\mathrm{script}},
  \label{eq:composition}
\end{equation}
where $\boldsymbol{h}_{i} \in \mathbb{R}^{(L+1)\times d}$ collects the residual
streams of position $i$ over all sites and each direction is non-zero only at its own site. Assuming $\alpha$ has already been set by the task grid search, we perform a second grid search to find the right site $\ell^{*}$ and amplitude $\beta$ for the script-normalization steering.  

\begin{table*}[t]
  \centering
  \caption{Task steering results on ASR, ER and AC. Rows with a prompt are averaged over three prompts for each task. Best score for each system is reported in bold. Since our focus is solely on training-free methods, we report LoRA-SFT in grey.}
  \label{tab:main}
  \small
  \setlength{\tabcolsep}{4.2pt}
  \begin{tabular}{ccc | *{4}{c} | *{5}{c} | c | *{2}{c}}
    \toprule
    \multicolumn{3}{c|}{\textbf{Method}}
    & \multicolumn{4}{c|}{\textbf{LibriSpeech }(WER$\downarrow$)}
    & \multicolumn{5}{c|}{\textbf{CommonVoice }(CER$\downarrow$)}
    & \textbf{IEMOCAP}
    & \multicolumn{2}{c}{\textbf{AudioCaps}} \\
    \cmidrule(lr){1-3}\cmidrule(lr){4-7}\cmidrule(lr){8-12}\cmidrule(lr){13-13}\cmidrule(lr){14-15}
    \makecell{Prompt} & \makecell{Task\\Steering} & \makecell{LoRA\\SFT}
    & \makecell{\textit{dev}\\\textit{clean}} & \makecell{\textit{dev}\\\textit{other}} & \makecell{\textit{test}\\\textit{clean}} & \makecell{\textit{test}\\\textit{other}}
    & \textit{en} & \textit{fr} & \textit{ar} & \textit{hi} & \textit{ru}
    & WA$\uparrow$
    & R-1$\uparrow$ & CIDEr$\uparrow$ \\
    \midrule
    \rowcolor{gray!15}
    \multicolumn{15}{l}{\textit{Qwen2-Audio-7B-Instruct}} \vspace{1mm} \\
    \cmark & \xmark & \xmark
      & 2.77 & 5.33 & 2.94 & 5.66
      & 7.4 & 9.1 & 52.3 & 51.7 & 22.7
      & 70.0 & 34.1 & 0.26 \\
    \xmark & \cmark & \xmark
      & 6.58 & 10.6 & 6.03 & 10.2
      & 13.2 & 10.8 & 68.1 & 73.3 & 32.0
      & 63.2 & \textbf{40.1} & \textbf{0.43} \\
    \cmark & \cmark & \xmark
      & \textbf{2.43} & \textbf{4.87} & \textbf{2.52} & \textbf{5.16}
      & \textbf{6.9} & \textbf{7.9} & \textbf{45.9} & \textbf{46.6} & \textbf{21.3}
      & \textbf{74.2} & 36.0 & 0.30 \\
    \baselinesteersep
    \xmark & \xmark & \cmark
      & \textcolor{gray}{7.54} & \textcolor{gray}{10.0} & \textcolor{gray}{8.14} & \textcolor{gray}{10.1}
      & \textcolor{gray}{7.1} & \textcolor{gray}{6.1} & \textcolor{gray}{30.2} & \textcolor{gray}{40.7} & \textcolor{gray}{14.4}
      & \textcolor{gray}{75.3} & \textcolor{gray}{48.2} & \textcolor{gray}{0.64} \\
    \midrule
    \rowcolor{gray!15}
    \multicolumn{15}{l}{\textit{Phi-4-Multimodal-5.6B-Instruct}} \vspace{1mm} \\
    \cmark & \xmark & \xmark
      & 2.12 & 4.16 & 2.26 & 4.71
      & 8.4 & 20.5 & 111.2 & 47.8 & \textbf{91.1}
      & 44.0 & 31.8 & \textbf{0.18} \\
    \xmark & \cmark & \xmark
      & 1.90 & 4.55 & 2.11 & 5.24
      & 11.1 & 12.2 & 101.4 & 72.4 & 179.8
      & 41.2 & 23.7 & 0.06 \\
    \cmark & \cmark & \xmark
      & \textbf{1.73} & \textbf{3.68} & \textbf{1.78} & \textbf{4.07}
      & \textbf{5.2} & \textbf{3.6} & \textbf{100.6} & \textbf{37.9} & 91.4
      & \textbf{45.8} & \textbf{32.6} & 0.17 \\
    \baselinesteersep
    \xmark & \xmark & \cmark
      & \textcolor{gray}{4.66} & \textcolor{gray}{7.81} & \textcolor{gray}{4.28} & \textcolor{gray}{7.25}
      & \textcolor{gray}{6.4} & \textcolor{gray}{5.8} & \textcolor{gray}{104.7} & \textcolor{gray}{68.7} & \textcolor{gray}{92.1}
      & \textcolor{gray}{43.9} & \textcolor{gray}{39.2} & \textcolor{gray}{0.33} \\
    \bottomrule
  \end{tabular}
  \vspace{-1em}
\end{table*}

\section{Experiments}
\label{sec:experiments}

We evaluate our steering methodology across two instruction-following SpeechLLMs, \textit{Qwen2-Audio-7B-Instruct}~\cite{qwen2audio} and \textit{Phi-4-Multimodal-5.6B-Instruct}~\cite{phi4mm}, on Automatic Speech Recognition (ASR), Emotion Recognition (ER) and Audio Captioning (AC). Script-Normalization (SN) corrections are solely applied in the case of ASR. For all tasks, directions are extracted from training data and evaluated on each dataset's official test set. For ASR, we gather 50 utterances from LibriSpeech \textit{train-100h}~\cite{LibriSpeech} and 50 utterances per language from the CommonVoice~\cite{commonvoice} training sets of English, French, Arabic, Hindi and Russian, for a total of 300 utterances. We extract the transcription direction on this small assembled dataset, and report Word Error Rate (WER) on LibriSpeech and Character Error Rate (CER) on CommonVoice. 
We further extend our protocol to ER using the IEMOCAP~\cite{iemocap} dataset, following the same evaluation procedure as~\cite{emo2vec}: a leave-one-session-out five-fold cross-validation with 20\% of each training split held out for validation. We subsequently extract the direction from 50 utterances of each training session, (sampled as evenly as possible across emotions) and evaluate on the held-out session, reporting Weighted Accuracy (WA). In our three prompts, we provide the four possible emotions to choose from: \textit{angry, happy, neutral, sad}. We also evaluate AC on AudioCaps~\cite{audiocaps}, extracting the direction from 100 random training clips and scoring with Consensus-based Image Description Evaluation (CIDEr)~\cite{cider}, which scores $n$-gram overlap with the reference caption while downweighting generic phrases, and ROUGE-1~\cite{rouge}, which measures unigram overlap between the generated caption and the reference. 
Finally, for SN, we do not require any training data. We simply extract the script direction by prompting the LLM to translate a textual english sentence to the target language, for 30 sentences (as described in section \ref{sec:method}). For each SpeechLLM, we manually verify that the meaning of each translated output matches the source sentence. As we want to evaluate the impact of steering along with prompting, performance may heavily vary with the choice of prompt, therefore confounding the effect of CAA. To reduce variability and show the generalizability of our approach across prompts, we score all our prompting results (Tables~\ref{tab:main}, ~\ref{tab:translation} and ~\ref{tab:cross_vs_full}) on the mean of three textual prompts. For transcription on non-latin scripts, we specifically provide the target script for each prompt, (e.g.\ for Russian: \textit{Transcribe the speech word for word in Cyrillic script. Output only the transcription.}) to make sure that the SpeechLLM has information about the target language.

\vspace{1mm}

Since our approach is training-free, we find insightful to compare our methodology against Supervised Fine-Tuning (SFT). For SFT, we train each task using Low-Rank Adaptation (LoRA)~\cite{lora} adapters ($r{=}16$, $\alpha{=}32$, dropout $0.05$) on the LLM attention projections only, feeding the same datasets as the ones used to extract the direction of each task. As SFT requires a development set, we apply early stopping on the lowest development-set loss for each task. Because we focus on inducing speech tasks in SpeechLLMs, all fine-tuning runs are done by feeding the audio only, which explains the discrepancy in results compared to prompting the model. Furthermore, since each fine-tuning run sees so few audio examples (between 50 and 300 utterances depending on the task), we observe strong overfitting during training, which likely explains the lack of improvement on some datasets, such as LibriSpeech (see Table \ref{tab:main}).



\section{Results}
\label{sec:results}

\subsection{Activation sites and direction injection}

\begin{figure}[t]
  \centering
  \captionsetup[subfigure]{labelformat=empty, skip=2.5pt}
  \subcaptionbox{\hspace*{-0.07\linewidth}%
      \shortstack{\textbf{Automatic Speech Recognition}\\
                  {\footnotesize (LibriSpeech + CommonVoice)}}%
      \label{fig:heatmap-asr}}%
    [\dimexpr0.5\columnwidth-2pt\relax]{%
      \includegraphics[width=\linewidth]{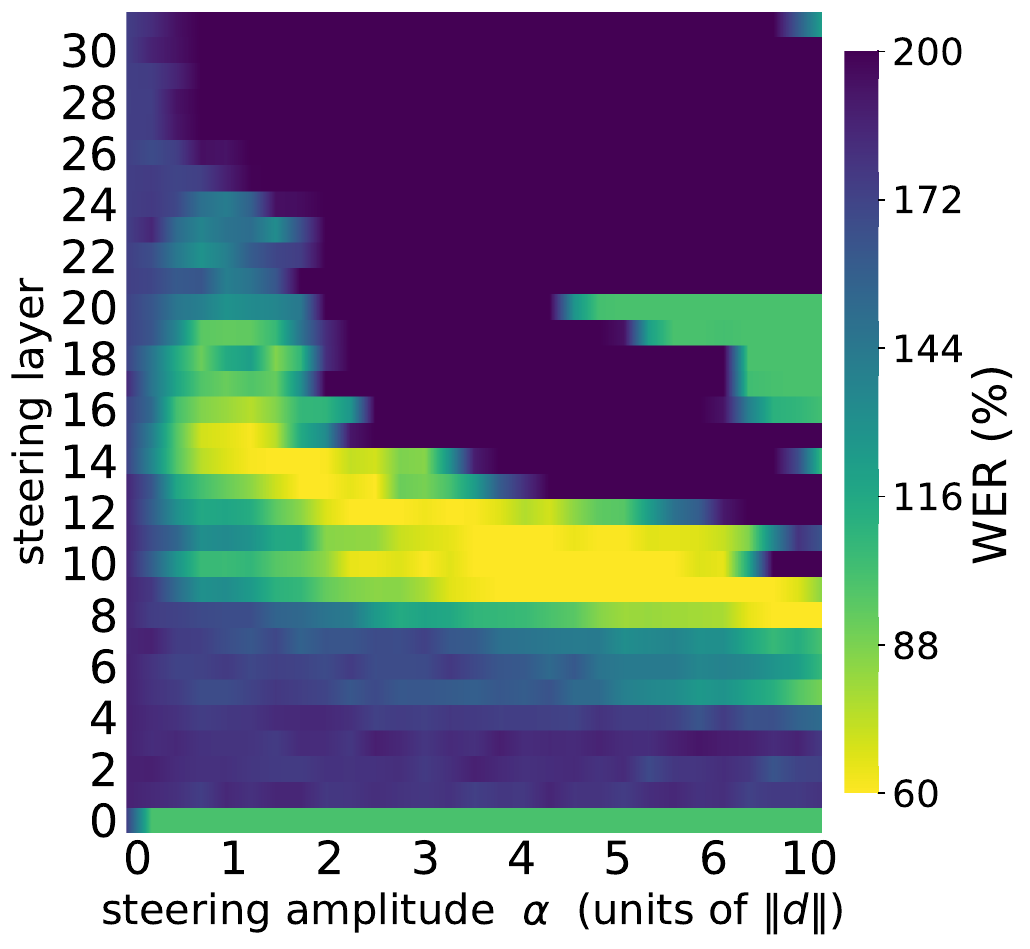}}%
  \hfill
  \subcaptionbox{\hspace*{-0.10\linewidth}%
      \shortstack{\textbf{Emotion Recognition}\\
                  {\footnotesize (IEMOCAP)}}%
      \label{fig:heatmap-er}}%
    [\dimexpr0.5\columnwidth-2pt\relax]{%
      \includegraphics[width=\linewidth]{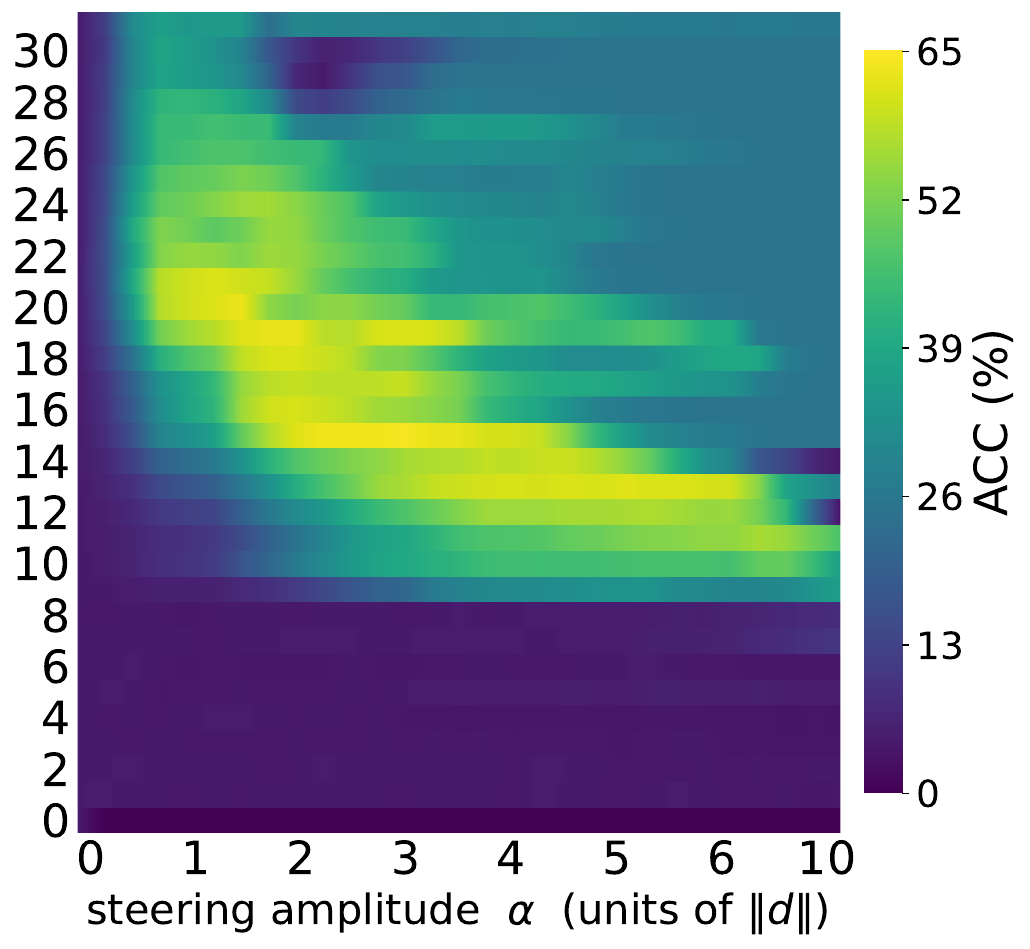}}
  \caption{grid searches for task induction through activation addition (w/o prompt), on \textit{Qwen2-Audio-7B-Instruct}, using the training set assembled in Section~\ref{sec:experiments}.}
  \label{fig:heatmaps}
\end{figure}

Figure~\ref{fig:heatmaps} shows a visual example of the $(\ell, \alpha)$ grid searches performed on ASR and ER. Effective injection sites concentrate in the middle layers (10 to 16) for both tasks, displaying a clustered pattern with progressively lower intensities towards the edges. The need for lower $\alpha$ at deeper layers follows from the Add\&Norm structure of the residual stream. An injection at layer~$\ell$ persists into all later blocks but is renormalized by fewer of them. Therefore, deeper injection sites achieve the same effect with a smaller amplitude and become over-saturated at larger values, which can prevent the SpeechLLM from carrying out the targeted task.

\vspace{1mm}

Having selected the best injection site and amplitude, we compare injecting the task steering vector on text positions only, against steering all positions (as depicted in Figure \ref{fig:method}).
Table~\ref{tab:taskonly_textonly} reports WER and CER results on LibriSpeech and CommonVoice for both methods. As observed, steering both text and audio tokens yields substantially better transcription than steering text tokens only ($43.9\%$ and $19.1\%$ relative on each dataset). Although the intervention shifts the audio activations, the acoustic content remains mostly intact and the behavior is more strongly elicited. A similar trend was observed on \textit{Phi-4-Multimodal-5.6B-Instruct}.

\vspace{1mm}

\begin{table}[H]
  \centering
  \caption{Impact of task steering over the input prompt tokens, on ASR using \textit{Qwen2-Audio-7B-Instruct}.
    LibriSpeech WER is the mean of the four splits dev-c/o and test-c/o. CommonVoice CER is the mean over the en/fr/ar/hi/ru test sets.}
  \setlength{\tabcolsep}{4pt}
  \begin{tabular}{l | c | c}
    \toprule
    \textbf{Task vector} & \textbf{LibriSpeech} & \textbf{CommonVoice} \\[-1mm]
    \cmidrule(lr){2-2}\cmidrule(lr){3-3}
    \textit{injected at}
      & WER$\downarrow$
      & CER$\downarrow$ \\
    \midrule
    Text tokens
      & 14.90 & 48.8 \\
    Text  + Audio tokens
      & \textbf{8.35} & \textbf{39.5} \\
    \bottomrule
  \end{tabular}
  \label{tab:taskonly_textonly}
\end{table}

\subsection{Contrastive Activation Addition results}

\newcommand{\dd}[1]{\,{\scriptsize\textcolor{gray}{(#1)}}}

Table~\ref{tab:main} reports CAA results for ASR, ER, and AC on both \textit{Qwen2-Audio-7B-Instruct} and \textit{Phi-4-Multimodal-5.6B-Instruct}, across task-induction methods (prompting, steering, and SFT). Note that we only evaluate the instruction-tuned versions of each model because they allow free-form prompting, unlike the base versions~\cite{qwen2audio}; this explains why our LibriSpeech WERs are slightly higher than results reported in technical reports of base models.

\vspace{1mm}

Applying task steering alone shows that inducing task behaviors into a SpeechLLM without textual instructions is indeed possible. On LibriSpeech, we achieve competitive WER results with a direct LoRA-SFT in the case of Qwen2-Audio, and better results than prompting for Phi-4, on the clean sets. On captioning, steering alone strongly improves CIDEr and ROUGE-1 over prompting on Qwen2-Audio ($0.26 \rightarrow 0.43$ and $34.1 \rightarrow 40.1$), but not on Phi-4-Multimodal. We justify this result by the fact that Qwen2-Audio has a default tendency to describe a given audio when no instruction is given, which in conjunction with steering, helps promote the captioning task. On the more challenging CommonVoice dataset, while Phi-4 shows promising results on French, overall CER scores are more mixed for both models, particularly on non-Latin-script languages (Arabic, Hindi, Russian).

\vspace{1mm}

Manual inspection of high WER and CER transcripts reveals \textbf{two} recurring failure modes.
First, the model often transcribes the utterance correctly, but fails to stop generating new tokens, therefore hallucinating a continuation. This means that, while $\boldsymbol{r}_{\mathrm{task}}$ elicits the
transcription behavior, it does not by itself encode when the audio has ended.
Second, on non-Latin-script languages, notably Russian and Arabic, many hypotheses preserve the semantic content of the target reference, yet are written in Latin script, implying that WER/CER penalizes script choice rather genuinely wrong transcription. A similar problem is observed when prompting the model without steering, even though the target script was specified in the instruction (see Section~\ref{sec:experiments}).

  

\vspace{1mm}

Adding the task vector on top of the prompt outperforms prompting alone on every benchmark for Qwen2-Audio, and on all but two (CommonVoice Russian and AudioCaps CIDEr) for Phi-4-Multimodal, with the largest absolute gains on CommonVoice. On Qwen2-Audio, this method lowers CommonVoice CER by 5.1 and 6.4 points absolute on Hindi and Arabic, while also improving consistently over the already low WER of LibriSpeech. On Phi-4-Multimodal, gains are largest on LibriSpeech test sets (21.2\% relative improvement on test-clean, 13.6\% on test-other) and on CommonVoice French and English (82.4\% and 38.1\% relative CER reduction). On Phi-4-Multimodal, we observe poor performance on Arabic and Hindi, consistent with the lack of support for these languages by the model.
On ER, Qwen2-Audio raises WA by 4.2 points, producing better overall emotion predictions and approaching SFT performance, while gains are more modest on Phi-4-Multimodal.
Inspecting the error reductions, combining steering with prompting improves ASR, ER and AC in several ways. We observe improved wording and syntax in the generated output, while also reducing cases where the instruct model simply refuses or fails to understand the required task, particularly on CommonVoice. It also prevents the model from transcribing or captioning past the end of the audio, a failure mode observed with task steering alone.

\vspace{1mm}

Without any prompt, steering outperforms LoRA-SFT on LibriSpeech for both SpeechLLMs, making CAA a viable alternative to fine-tuning when labeled data is scarce. On the more challenging CommonVoice, however, SFT remains the stronger option.

\vspace{1mm}

\begin{table}[H]
  \centering
  \caption{Script-Normalization direction combined with Prompt and Task Steering for CommonVoice on non-Latin scripts (Arabic, Hindi and Russian). CER is reported.}
  \label{tab:translation}
  \small
  \setlength{\tabcolsep}{3.5pt}
  \begin{tabular}{@{}l | ccc | ccc@{}}
    \toprule
    & \multicolumn{3}{c|}{\textbf{Qwen2-Audio}} & \multicolumn{3}{c}{\textbf{Phi-4-MM}} \\
    \textbf{Method} & ar & hi & ru & ar & hi & ru \\
    \midrule
    Prompt \& Task Steering
      & 45.9 & 46.6 & 21.3 & \textbf{100.6} & 37.9 & 91.4 \\
      \; \ + Script Steering
      & \textbf{40.5} & \textbf{44.1} & \textbf{21.2} & 102.4 & \textbf{36.7} & \textbf{62.2} \\
    \bottomrule
  \end{tabular}
\end{table}

\vspace{1mm}

Given the limitations of steering on non-Latin-script languages, Table~\ref{tab:translation} applies the SN direction, in an attempt to steer the model towards a targeted alphabet. Results show clear improvements for some languages: on Phi-4-Multimodal, CER drops by 29.2 points on Russian, largely by correcting the script, and by 1.2 points on Hindi, while Arabic stagnates; on Qwen2-Audio, CER drops by 5.4 points on Arabic and 2.5 points on Hindi. 

\subsection{Cross-domain transfer and steering robustness}


From the 300-utterance ASR training set used to extract the transcription direction, we isolate utterances from each dataset (LibriSpeech and CommonVoice). We evaluate each direction both across domains (e.g., CommonVoice $\rightarrow$ LibriSpeech) and within the same domain (e.g., LibriSpeech $\rightarrow$ LibriSpeech). All experiments use the same $(\ell,\alpha)$ pairs as in Table~\ref{tab:main}.

\begin{table}[h]
  \centering
  \small
  \setlength{\tabcolsep}{3pt}
  \caption{Out-of-domain transfer on \qwen{} for ASR (reported on the same sets as in Table \ref{tab:taskonly_textonly}). The $\pm$ values in the \textbf{Prompt} columns give the standard deviation over the three prompts. \textit{Random} replaces the task vector with a random one of equal norm at the same site.}
\vspace{1mm}
  \label{tab:cross_vs_full}
  \begin{tabular}{c | cc | cc}
    \toprule
    \textbf{Task vector} & \multicolumn{2}{c|}{\textbf{w/o Prompt}} & \multicolumn{2}{c}{\textbf{Prompt}} \\[-1mm]
    \cmidrule(lr){2-3} \cmidrule(lr){4-5}
    \textit{extracted from}
      & LS\,WER$\downarrow$ & CV\,CER$\downarrow$
      & LS\,WER$\downarrow$ & CV\,CER$\downarrow$ \\
    \midrule
    LS\,+\,CV
      & 8.35 & 39.5 & 3.75\s{0.19} & 25.7\s{1.4} \\
    LS
      & \cellcolor{gray!15}\textbf{8.31} & \textbf{38.5}
      & \cellcolor{gray!15}\textbf{3.75}\s{0.18} & 25.6\s{1.3} \\
    CV
      & 8.58 & \cellcolor{gray!15}38.7
      & 3.76\s{0.17} & \cellcolor{gray!15}\textbf{25.5}\s{1.2} \\
    \midrule
    No Steering
      & 239.9 & 159.7
      & 4.18\s{0.49} & 28.6\s{2.1} \\
    Random
      & 196.7 & 178.1
      & 4.31\s{0.54} & 29.7\s{3.0} \\
    \bottomrule
  \end{tabular}
\end{table}

Table~\ref{tab:cross_vs_full} shows a negligible performance drop when a direction extracted from LibriSpeech (LS) is applied to CommonVoice (CV), and vice versa, indicating that the transcription direction transfers across domains and is largely dataset-agnostic. Because the differences in observed WER/CER between directions are small, we include two negative controls to confirm that the gains come from the steering direction itself: no steering, and a random vector of equal norm injected at the same site. Without a prompt, both controls fail (WER/CER above 150\%), because the model receives no task specification and the random vector does not supply one. With a prompt, task steering consistently improves over prompting alone (LS WER $4.18 \rightarrow 3.75$, CV CER $28.6 \rightarrow 25.7$), whereas a random direction degrades performance below the prompt-only baseline, confirming that the gains are specific to the task direction. Task steering also reduces the standard deviation across prompts by roughly a factor of two. This suggests that steering makes transcription quality less sensitive to the choice of prompt.

\section{Conclusion}
\label{sec:conclusion}

We presented a training-free and low-resource procedure for extracting and inducing task-specific behaviors in SpeechLLMs via CAA, showing that speech tasks are linearly encoded in their representation space. Despite a limited-data regime, our protocol improves over prompting alone on nearly all ASR, ER and AC benchmarks for two distinct SpeechLLMs, while reducing sensitivity to prompt wording, and offers an alternative to low-resource SFT when no textual instruction is used. We further demonstrate that task steering vectors are localized and transferable to out-of-domain datasets, with a negligible impact on WER/CER for ASR. Similarly to textual LLMs, we discover that CAA applied to input audio embeddings does not affect the semantic of the given input when the right steering vector is applied, yielding to better overall steering performance. Finally, a script-normalization direction extracted from text alone, requiring no speech data, can be used to enforce a target alphabet on non-Latin-script languages, reducing CER (in the case of ASR on non-latin languages) while preserving the semantic content of the transcription.

\newpage

\section{Acknowledgments}
Severin Baroudi and Ricard Marxer were supported by the Agence de l'Innovation de Défense (grant 2022~65~0079), the French National Research Agency (ANR-20-CE23-0012-01, MIM), and computational resources from GENCI-IDRIS (grants AD011014044R2 and AD011016519), while Idiap employees were funded by the European Union's Horizon 2020 project ELOQUENCE (grant 101070558). This work has also been carried out while Pierfrancesco Melucci was enrolled in the Italian National Doctorate on Artificial Intelligence run by Sapienza University of Rome in collaboration with Universita di Roma Tre, and Severin Baroudi was researcher at the Mila -- Quebec AI Institue. 

\section{Compliance with Ethical Standards}
This research was conducted retrospectively using human subject speech data made available in open access or under institutional license (LibriSpeech, CommonVoice, IEMOCAP, AudioCaps). Ethical approval was not required, as confirmed by the licenses of these corpora. No new data were collected from human subjects.

\bibliographystyle{IEEEbib}
\bibliography{refs}

\end{document}